\documentclass{article}

\usepackage[final]{nips_2018}

\usepackage[utf8]{inputenc} 
\usepackage[T1]{fontenc}    
\usepackage{hyperref}       
\usepackage{url}            
\usepackage{booktabs}       
\usepackage{amsfonts}       
\usepackage{nicefrac}       
\usepackage{microtype}      
\usepackage{amsmath}
\usepackage{mathtools}
\usepackage{amsthm}
\usepackage{thmtools}
\usepackage{amssymb}
\usepackage{tikz,tikz-cd}
\usepackage{wrapfig}

\DeclareMathOperator*{\argmin}{arg\,min}

\newtheorem{thm}{Theorem}
\newtheorem{lem}[thm]{Lemma}
\newtheorem{cor}[thm]{Corollary}
\newtheorem{rem}[thm]{Remark}

\declaretheoremstyle[%
  spaceabove=6pt,%
  spacebelow=6pt,%
  headfont=\normalfont\itshape,%
  postheadspace=0.5em,%
  qed=\qedsymbol%
]{mystyle}
\declaretheorem[name={Proof},style=mystyle,unnumbered,
]{prf}

\newcommand{\M}{\mathcal{M}}
\newcommand{\X}{\mathcal{X}}
\newcommand{\Y}{\mathcal{Y}}
\newcommand{\Z}{\mathcal{Z}}
\newcommand{\R}{\mathbb{R}}
\newcommand{\id}{\text{id}}
\newcommand{\vv}{\mathbf{v}}
\newcommand{\uu}{\mathbf{u}}
\newcommand{\x}{\mathbf{x}}
\newcommand{\y}{\mathbf{y}}
\title{Topological Constraints on\\Homeomorphic Auto-Encoding}

\author{
  Pim de Haan\thanks{Equal contribution}\\
  University of Amsterdam, UC Berkeley \\
  \texttt{pimdehaan@gmail.com}
  \And
  Luca Falorsi\footnotemark[1] \\
  University of Amsterdam \\
  \texttt{luca.falorsi@gmail.com}
}

\begin{document}

\maketitle

\begin{abstract}
When doing representation learning on data that lives on a known non-trivial manifold embedded in a high dimensional space, it is natural to desire the encoder to be \emph{homeomorphic} when restricted to the manifold, so that it is bijective and continuous with a continuous inverse.
Using topological arguments, we show that when the manifold is non-trivial, the encoder must be globally discontinuous and propose a universal, albeit impractical, construction. In addition we derive necessary constraints which need to be satisfied when designing manifold-specific practical encoders.
These are used to analyse candidates for a homeomorphic encoder for the manifold of 3D rotations $SO(3)$.
\end{abstract}

\section{Introduction}

Unsupervised learning attempts to learn to encode data living in a high dimensional space in a more useful low dimensional representation.
This is often reasonable assuming the manifold hypothesis, which states that real world high dimensional data lives on a low dimensional manifold.

Popular methods methods for representation learning, such as Auto-Encoders, learn an encoder which represents the data in Euclidean spaces as well as a decoder that reconstructs the original representation from the latent space. However, when the true manifold of the data has a non-trivial topology, encoding the data in a Euclidean space leads to poor representations.

In such scenarios, one of two things can happen. When the latent space is sufficiently bigger than the intrinsic dimensionality of the true manifold, the true manifold is again embedded in the latent space, which leads to low density areas in the latent space. In the other case, when the dimensionality of the Euclidean latent space is comparable to the dimensionality of the true manifold, but the true manifold can not be embedded in the latent space, the encoder must be discontinuous. This problem was described in \citet{davidson2018hyperspherical} and \citet{falorsi2018explorations}

In many situations, the true manifold is a known non-trivial manifold, for example, when inferring the pose of an object from images in an unsupervised fashion or when inferring joint angles of images of a robot arm. Three natural requirements of an ideal auto encoder can then be formulated. Firstly, we desire the encoder to be bijective, so that every element of the latent space corresponds to exactly one point on the data manifold. Secondly, we desire continuous trajectories on the data manifold, such as motion of robot joint angles, to correspond to continuous trajectories in the latent space. Last, we desire conversely that continuous paths in the latent space correspond to continuous paths on the data manifold, which is to say that latent space interpolations are correct. An encoder that satisfies these requirements is referred to in topology as a \emph{homeomorphism}. When combined with a suitable decoder, we call it a \emph{Homeomorphic Auto-Encoder}. However, the concept translates to popular representation learning methods like the Variational Auto-Encoder, BiGAN and InfoGAN as well.

One immediate consequence of having a homeomorphic encoder is that the latent space is homeomorphic to the true data manifold. This rules out Euclidean spaces for non-trivial true data manifolds. Since neural networks only map to Euclidean spaces, they must be combined with specifically designed functions. Topological arguments allows us to derive necessary and sufficient conditions for such a construction to be a homeomorphic Auto-Encoder.

\section{The encoder is globally discontinuous}
We assume the data lives on a path connected manifold
$\M$ topologically embedded in the data space $\X := \R^n$ for some $n$ with embedding $i : \M \hookrightarrow \X$. The image of the embedding is submanifold $\X' := i(\M)\subseteq \X$. We desire encoder $\psi : \X \to \M$ to be such that its restriction $\psi' := \psi|_{\X'}$ is a homeomorphism.
By the following theorem, this immediately leads to a somewhat surprising observation: when $\M$ is non-trivial, the encoder must be \emph{globally} discontinuous, even though it is continuous when restricted to $\X'$.


\begin{thm}\label{thm:retract}
Let $\psi$ be defined as above. Assume additionally that $\psi$ is continuous, in which case $\psi$ is a retract. Then for all $k$, the $k$-th homotopy group of $\M$, $\pi_k(\M)$, is isomorphic to a subgroup of $\pi_k(\X)$.
\end{thm}
\begin{prf}
$i$ and $\psi$ induce $k$-th homotopy group homomorphisms: $\psi_{*}:\pi_{k}(\X) \rightarrow\pi_{k}(\M)$ and $i_{*}:\pi_{k}(\M) \rightarrow\pi_{k}(\X)$.
By construction, the function $\psi\circ i: \M\to \M$ is a homeomorphism, so $(\psi \circ i)_{*}=\psi_{*}\circ{i_{*}}:\pi_{k}(\M) \rightarrow\pi_{k}(\M)$ is a group automorphism. Thus $i_{*}$ must be injective, which implies the thesis.
\end{prf}
\begin{cor}\label{cor:discontinuous}
Let $\psi, \M$ be defined as above. Assume $\M$ having a non-trivial $k$-th homotopy group for some $k$. Then no such $\psi$ exist that is additionally continuous.
\end{cor}


\section{Existence of the encoder}\label{lem:suff-1}
We now turn to the question of whether a homeomorphic auto-encoder can always be constructed. Since Neural Networks only map Euclidean spaces to Euclidean spaces, we decompose our encoder $\psi = \phi \circ \xi$, for $\xi : \X \to \Y$ to be learned by a Neural Network, with $\Y$ a Euclidean space, and a designed function $\phi : \Y \to \M$. In the following subsections, we discuss the existence of $\phi$ and $\xi$.

\subsection{Existence and continuity of the map $\Y \to \M$}

Assuming the dimensionality of $\Y$ is sufficiently large, an embedding can be constructed for any manifold $\M$, by the Whitney Embedding Theorem (Theorem 2.6 in \citep{adachi1993embeddings}), which states that any manifold of $d$ dimensions can be embedded in $\R^{2d+1}$ as a closed subset. 
We take this Euclidean space to be $\Y$ and call the closed embedding $j : \M \to \Y$. 

Then the metric projection can be defined in the following way.

For any $x \in \Y$, we can define the distance to $j(\M)$:
\begin{align*}
    d_{j(\M)}(x) := \inf_{y \in j(\M)} \|x-y\| = \min_{y \in j(\M)} \|x-y\|
\end{align*}
where the last equality holds since $j(\M)$ is closed. Then the metric projection is:
\begin{align*}
    \pi_{j(\M)}(x) = \argmin_{y \in j(\M)}\|x-y\|
\end{align*}

The projection map may have multiple value may be discontinuous, but the following lemma shows that it is uniquely defined almost everywhere in $\Y$ and that the projection map is continuous almost everywhere.

Thus it is a potential candidate for the desired mapping.

\begin{lem}
For a closed set $C \subseteq \R^n$, for almost all points $x \in \R^n$, there is a unique point $y \in C$, such that $d_C(x)=\|x-y\|$. Additionally, the metric projection is continuous almost everywhere.
\begin{prf}
First, we show that $d_C$ is 1-Lipschitz. Let $x,y \in \R^n$. Without loss of generality, choose $d_C(x) \ge d_C(y)$ and let $w \in C$ s.t. $d_C(y)=\|y-w\|$. Then:
\begin{align*}
    | d_C(x) - d_C(y) | = \inf_{z \in C} \|x- z\| - \|y- w\| \le \|x- w\| - \|y- w\| \le \|x- y\|
\end{align*}
where in the last step we used the triangle inequality.

Therefore, by Rademacher's theorem \citep{herbertfederer1996}, $d_C$ is differentiable almost everywhere.

The envelope theorem in \cite{milgrom2002envelope} states that if $d_C$ is differentiable in $x \in \R^n$ and $y\in C$ s.t. $d_C(x)=d(x,y)$, then $\nabla d_C(x) = \nabla_x d(x, y)$, so that:
$$\nabla d_C(x) = \nabla_x \|x - y \| = \frac{x-y}{\|x-y\|}$$

From this expression of the gradient, we can derive uniqueness of the minimising point. Let $d_C$ be differentiable at $x$ and let $y$ and $y'$ have that $d_C(x) =\|x-y\|=\|x-y'\|$, we have:
\[
\nabla d_C(x) = \frac{x-y}{\|x-y\|}=\frac{x-y'}{\|x-y'\|} \implies y=y'
\]
Furthermore, we have, for all $x$ such that $d_C$ is differentiable:
\[ \pi_C(x)=x-d_C(x)\nabla d_C(x) \]
so that $\pi_C$ is continuous almost everywhere.
\end{prf}
\end{lem}

The case of the hypersphere $S^n$ nicely illustrates these properties. The manifold can be embedded in $\R^{n+1}$ and the projection map is the normalisation: $\phi : \R^{m+1} \to S^n : \x \mapsto \x / \| \x \|$. This map is well-defined and continuous everywhere, except on the set containing just the origin, which is of measure 0. In this particular case the projection map is easily computable and smooth almost everywhere, but this is not the case for any manifold.

\subsection{Existence and continuity of the map $\X \to \Y$}
By the above argument we can construct an embedding $j : \M \to \Y$ and a map $\phi : \Y \to \M$ such that $\phi \circ j = \text{id}_\M$ and we assumed an embedding $i: \M \to \X$ exists.
Then the function $\xi' := j \circ i^{-1} : i(\M) \to j(\M)$, is a homeormorphism.

Since both $\X$ and $\Y$ are Euclidean, by the Tietze extention theorem \citep{munkres2000topology}, this map can be extended from the closed set $\X'$ to the entire space $\X$, making a continuous map $\xi : \X \to \Y$.
Since neural networks can approximate continuous functions arbitrarily well \citep{hornik1991approximation}, we can then try to learn a neural network to approximate $\xi$.

\section{Constraints on a practical encoder}
We can therefore always construct a homeormorphic auto-encoder $\psi: \X \to \M$ that is continuous almost everywhere, but the projection map may not be differentiable, nor easily computable, in which cases it is impractical for the learning the neural network. For many manifolds $\M$ a practical alternative can instead be hand-crafted. For such a construction we can derive several necessary conditions.

\subsection{Decomposing the encoder}
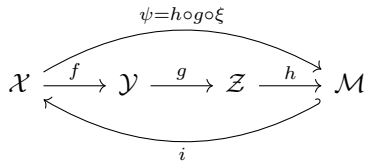
\begin{figure}[h!]
    \centering
    \begin{tikzcd}
           \X \arrow[r, "f"] \arrow[rrr, bend left=30, "\psi=h \circ g \circ \xi"] & \Y \arrow[r, "g"]  & \Z \arrow[r, "h"] & \arrow[lll, hook', "i", bend left=30 ]\mathcal{M}
    \end{tikzcd}
    \caption{Decomposition of the encoder. $\xi$ and $h$ are continuous. $g$ is generally discontinuous.}
    \label{fig:decomposition}
\end{figure}

Knowing that the encoder must be discontinuous for non-trivial manifolds, it can always be decomposed in the composition of a continuous function $\xi$, a discontinuous function $g$ and a continuous function $h$, where $\xi$ and $h$ are possibly identity maps. We can then define spaces as in Figure \ref{fig:decomposition}.

We additionally define subsets: $\X' := i(\X), \Y' := \xi(\X'), \Z' := g(\Y')$ and restricted maps $\xi' := \xi|_{\X'}, g' := g|_{\Y'},h' := h|_{\Z'}$ so that $\psi' = h' \circ g' \circ \xi'$.

\subsection{Necessary conditions}
When designing such an encoder, care has to be taken to ensure that the intermediate spaces and functions are such that it is possible to express a homeomorphic encoder. This leads to the following necessary conditions.

\begin{lem}\label{lem:nec-1}
Let $\xi$, $\xi'$, $\X$, $\Y$, $i$ and $\mathcal{M}$ be defined as above. If $\M$ is additionally compact and $\Y$ Hausdorff, then $\xi'$ is a homeomorphism and $\M$ is embeddable in $\Y$.
\begin{prf}
As defined above, $\xi$ is continuous and $\xi'$ bijective, so $\xi'$ is continuous bijective.
Since $\M$ compact then $\X'$ is also compact.
Let $C$ be any closed subset of $\X'$, then $C$ compact. Now, since continuous maps map compact sets to compact sets then $\xi'(C)$ is compact and, since $\Y'$ subspace of an Hausdorff space is Hausdorff, $\xi'(C)$ is closed. $\xi'$ is thus closed.
Then $\xi'$ is a homeomorphism and $\xi'\circ i$ is an embedding of $\M$ in $Y$
\end{prf}
\end{lem}

\begin{lem}\label{lem:nec-2}
Let $h$, $h'$, $\M$, $\Z$ and $\Z'$ be defined as above.
If additionally $\Z'$ is compact then $h'$ is a homeomorphism and $\M$ is embeddable in $\Z$.
\begin{prf}
As defined above, $h$ is continuous and $h'$ bijective, so $h'$ is continuous bijective.
Since $\Z'$ compact, $h'$ is closed, following an argument similar to the proof of Lemma \ref{lem:suff-1}, thus $h'$ is a a homeomorphism. $h'^{-1}$ is an embedding of $\mathcal{M}$ in $\Z$.
\end{prf}
\end{lem}
\begin{rem}
Let $h$, $h'$, $\M$, $\Z$ and $\Z'$ be defined as above.
Alternatively if we additionally assume $\Z$ is compact, then the continuity of $h$ implies that $\Z'$ is compact, since $\Z' = h^{-1}(\M)$ is closed and thus compact in $\Z$.
\end{rem}

Since we want to use neural networks to learn a homeomorphic mapping of $\X'$ to $\M$, we need to be able to be able to apply gradient descent methods. In order for this to be possible we need the discontinuity points of $g$ to be negligible (for example sets of measure $0$), or to add constraints on $\xi$ to encourage $\Y'$ to lie outside of the discontinuity regions. If we make this latter modelling choice, we then see that $\M$ needs to be embeddable in all intermediate spaces:
\begin{lem}\label{lem:all-homeo}
Let $\psi, \psi', \xi, \xi', g, g', h, h'$, $\X, \X', \Y, \Z, \Z'$ and $\M$ be as defined above.
If additionally $g'$ is assumed to be continuous, then $\xi'$, $g'$, $h'$ are all homeomorphisms and $\M$ must be embeddable in $\X$, $\Y$ and $\Z$.
\begin{prf}
Notice that by construction $\xi'$, $g'$, $h'$ are bijective functions.
We prove that $h'$ is open: Take $A$ open in $\Z'$, since $\xi'$, $g'$ are continuous then $\tilde A:= (\xi'\circ g')^{-1}(A)$ is open in $\X'$. Now since $\psi'$ is a homeomorphism $h'(A) = \psi'(\tilde A)\subseteq \X'$ is open.
Now using that $h'^{-1}\circ\psi$ is a homeomorphism we can prove that $g';$ is open.
Finally using that $g'^{-1}\circ h'^{-1}\circ\psi$ is a homeomorphism we can prove that $g';$ is open.
From this the thesis follows.
\end{prf}
\end{lem}
\begin{cor}\label{cor:homotopy-group}
Let $h, h'$, $\M$ and $\Z$ be defined as above.
If $h'$ additionally is a homeomorphism, by Theorem \ref{thm:retract} using the fact that $h$ is continuous, $h$ is a retract, so for all $k$, the $k$-th homotopy group of $\M$, $\pi_k(\M)$, is isomorphic to a subgroup of $\pi_k(\Z)$.
\end{cor}

\section{Application to $SO(3)$}
To show the utility of the proposed theory we it analyse four sensible candidate architectures for a homeomorphic encoder to the compact manifold of 3D rotations $SO(3)$. The candidates are taken from \cite{falorsi2018explorations}.
Throughout the next lines we will indicate with $\xi:\X\to \mathbb{R}^n$ a neural network and we will use the fact that $\pi_1(SO(3))= \mathbb{Z}/2\mathbb{Z}$.

\paragraph{Exponential map:}
$\xi:\X\to \mathbb{R}^3$, $g:=\id$, $h:=\exp:\mathbb{R}^3\to SO(3)$ \newline
where $\exp$ is the exponential map from the Lie algebra to the Lie group, which is always continuous and surjective for $SO(3)$. The encoder $h\circ g\circ \xi$ is globally continuous, this contradicts the necessary condition following from Theorem \ref{thm:retract}.

\paragraph{Quaternions:}
$\xi:\X\to \mathbb{R}^4$, $g:\mathbb{R}^4\to S^3,\ \x\mapsto \dfrac{\x}{\|\x\|}$, $h:S^3 \to SO(3)$
\newline
where $h$ is the surjective continuous map from quaternions to $SO(3)$.
Since $S^3$ is compact, then Lemma \ref{lem:nec-2} applies. Since $S^3$ is simply connected, this contradicts the necessary condition following from Corollary \ref{cor:homotopy-group}.

\paragraph{Axis-Angle\protect\footnote{Here, $v_1,v_2$ indicate the components of $S^1\subseteq \mathbb{R}^2$, and $[\cdot]_\times:\mathbb{R}^3\to \mathfrak{so(3)}$ the vector space isomorphism that maps $\mathbb{R}^3$ to the lie algebra of $SO(3)$, i.e. the space of skew-symmetric $3\times 3$ real matrices}:}
$\begin{aligned}
&\xi : \X \to \mathbb{R}^3 \times \mathbb{R}^2, \quad
g:\mathbb{R}^3\times\mathbb{R}^2\to S^2\times S^1,\ (\x,\y)\mapsto (\dfrac{\x}{\|\x\|},\dfrac{\y}{\|\y\|}) \\
&h:S^2\times S^1\to SO(3),\ (\uu,\vv)\mapsto I + v_1\cdot \uu_\times + v_2\cdot \uu_\times^2
\end{aligned}$
\newline
where $S^2\times S^1$ correspond to the axis and the angle of the rotation. Since $S^2\times S^1$ is compact, then by Lemma \ref{lem:nec-2}, $h'$ must be a homeomorphism, and thus $SO(3)$ embeddable $S^2\times S^1$, which is false. Thus the necessary conditions are not met.

\paragraph{Basis:}
$\begin{aligned}
& \xi:\X\to \mathbb{R}^3\times\mathbb{R}^3, \quad
g:\mathbb{R}^3\times\mathbb{R}^3\to S^2\times S^2,\ (\x,\y)\mapsto (\dfrac{\x}{\|\x\|},\dfrac{\y}{\|\y\|}), \\
& h:S^2\times S^2\to SO(3), (\mathbf{u}, \vv) \mapsto \text{concat}(\mathbf{w_1}, \mathbf{w_2}, \mathbf{w_3}), \\
&\mathbf{w_1} = \mathbf{u},\; \mathbf{w_2}' = \vv - \langle \mathbf{u},\vv \rangle \mathbf{u},\; \mathbf{w_2} = \frac{\mathbf{w_2}'}{\|\mathbf{w_2}'\|},\; \mathbf{w_3} = \mathbf{w_1}\times \mathbf{w_2}
\end{aligned}$
\newline
This follows from the fact that $SO(3)$ can be seen as the (oriented) bases of a three dimensional space. This construction satisfies the sufficient condition following from Lemma \ref{lem:suff-1}.
We see this by defining $\phi:= h\circ g: \R^3 \times \R^3 \to SO(3)$ and $\Y':= \{(\uu,\vv)\in \R^3 \times \R^3: \|\uu\|=\|\vv\|=1, \langle\uu,\vv \rangle = 0\}$, then $\phi$ satisfies the hypothesis of Lemma \ref{lem:suff-1}. In fact $\phi'$ is a homeomorphism since for the $SO(3)$ matrices, the last column is automatically determined by the first two.

Note that these theoretical results are consistent with the experimental results in \cite{falorsi2018explorations}, where it was found that only the `Basis' method learns a homeomorphic encoder.

\section{Conclusions and future work}
We developed a theoretical framework for the analysis of architectures of Auto-Encoders for non-trivial manifolds. Several necessary conditions were stated for the Auto-Encoder to be homeomorphic, as well as a sufficient condition, which can guide the development of homeomorphic Auto-Encoders for various manifolds. For the $SO(3)$ manifold it was found that the theoretical results were consistent with experimental results in prior work.

In future work we will try to generalise the developed  theory using other topological invariants besides homotopy and to situations where the manifold is not embedded, but imbedded or embedded with symmetries.
In addition we would like to analyse what happens when the embedding is noisy in a probabilistic framework.
Closer attention will be given to investigating to what extent these results are relevant for supervised problems, such as pose estimations.
Finally we will try to apply the principles of the homeomorphic Auto-Encoders to practical problems and existing architectures, such as a version of \cite{Eslami1204} without supervised pose labels.

\subsubsection*{Acknowledgments}
We would like to thank Patrick Forr\'e for his suggestions on improving an earlier draft and Tim Davidson and Taco Cohen for their helpful discussions.
\bibliographystyle{apalike}
\bibliography{main}

\end{document}